\documentclass{article}
\usepackage{utils/iclr2021_conference,times}

\usepackage{amsmath,amsfonts,bm}

\def\Figref#1{Figure~\ref{#1}}

\def\Secref#1{Section~\ref{#1}}

\def\eqref#1{equation~\ref{#1}}

\def\Apxref#1{Appendix~\ref{#1}}

\def\Tabref#1{Table~\ref{#1}}

\def\1{\bm{1}}

\def\rx{{\textnormal{x}}}

\def\rvx{{\mathbf{x}}}

\def\vx{{\bm{x}}}

\DeclareMathAlphabet{\mathsfit}{\encodingdefault}{\sfdefault}{m}{sl}
\SetMathAlphabet{\mathsfit}{bold}{\encodingdefault}{\sfdefault}{bx}{n}

\DeclareMathOperator*{\argmax}{arg\,max}

\usepackage{utils/packages}
\usepackage{utils/commands}

\title{Hyperparameter Transfer Across Developer Adjustments}

\author{
Danny Stoll${}^{1}$, Jörg K.H. Franke${}^{1}$, Diane Wagner${}^{1}$, Simon Selg${}^{1}$ \& Frank Hutter${}^{1,2}$\\
${}^1$University of Freiburg \\
${}^2$Bosch Center for Artificial Intelligence\\
\texttt{\{stolld,frankej,wagnerd,selgs,fh\}@cs.uni-freiburg.de}}

\iclrfinalcopy

\DeclareUnicodeCharacter{2212}{-}
\begin{document}

\maketitle

\begin{abstract}
After developer adjustments to a machine learning (ML) algorithm, how can the results of an old hyperparameter optimization (HPO) automatically be used to speedup a new HPO?
This question poses a challenging problem, as developer adjustments can change which hyperparameter settings perform well, or even the hyperparameter search space itself.
While many approaches exist that leverage knowledge obtained on previous \emph{tasks}, so far, knowledge from previous \emph{development steps} remains entirely untapped.
In this work, we remedy this situation and propose a new research framework: hyperparameter transfer across adjustments (HT-AA).
To lay a solid foundation for this research framework, we provide four simple HT-AA baseline algorithms and eight benchmarks
changing various aspects of ML algorithms, their hyperparameter search spaces, and the neural architectures used.
The best baseline, on average and depending on the budgets for the old and new HPO, reaches a given performance 1.2--2.6x faster than a prominent HPO algorithm without transfer.
As HPO is a crucial step in ML development but requires extensive computational resources, this speedup would lead to faster development cycles, lower costs, and reduced environmental impacts.
To make these benefits available to ML developers off-the-shelf and to facilitate future research on HT-AA, we provide python packages for our baselines and benchmarks.
\end{abstract}

\begin{figure}[H]
    \centering
    \includegraphics[width=\linewidth]{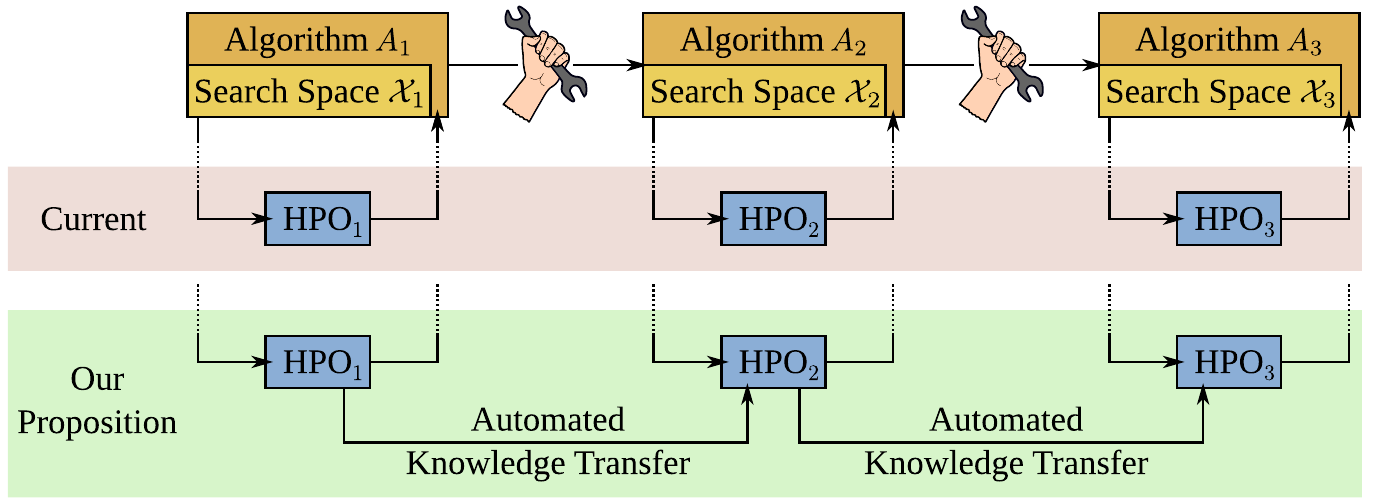}
    \caption*{Graphical Abstract:
        Hyperparameter optimization (HPO) across adjustments to the algorithm or hyperparameter search space.
        A common practice is to perform HPO from scratch after each adjustment or to somehow manually transfer knowledge.
        In contrast, we propose a new research framework about automatic knowledge transfers across adjustments for HPO.
    }
\label{fig:overview}
\end{figure}

\section{Introduction: A New Hyperparameter Transfer Framework}

The machine learning (ML) community arrived at the current generation of ML algorithms by performing many iterative adjustments.
Likely, the way to artificial general intelligence requires many more adjustments.
Each algorithm adjustment could change which settings of the algorithm's hyperparameters perform well, or even the hyperparameter search space itself \mbox{\citep{chen2018bayesian,Li2020Rethinking}}.
For example, when deep learning developers change the optimizer, the learning rate's optimal value likely changes, and the new optimizer may also introduce new hyperparameters.
Since ML algorithms are known to be very sensitive to their hyperparameters \citep{chen2018bayesian,feurer_hyperparameter_2019}, developers are faced with the question of how to adjust their hyperparameters after changing their code.
Assuming that the developers have results of one or several hyperparameter optimizations (HPOs) that were performed before the adjustments, they have two options:
\begin{itemize}
  \item[1.] Somehow manually transfer knowledge from old HPOs.
\end{itemize}
This is the option chosen by many researchers and developers, explicitly disclosed, e.g., in the seminal work on AlphaGo \citep{chen2018bayesian}.
However, this is not a satisfying option since manual decision making is time-consuming, often individually designed, and has already lead to reproducibility problems \citep{musgrave2020metric}.
\begin{itemize}
  \item[2.] Start the new HPO from scratch.
\end{itemize}
Leaving previous knowledge unutilized can lead to higher computational demands and worse performance (demonstrated empirically in \Secref{sec:experiments}).
This is especially bad as the energy consumption of ML algorithms is already recognized as an environmental problem.
For example, deep learning pipelines can have CO\textsubscript2 emissions on the order of magnitude of the emissions of multiple cars for a lifetime  \citep{strubell-etal-2019-energy}, and their energy demands are growing furiously: \citet{schwartz2019green} cite a ``300,000x increase from 2012 to 2018''.
Therefore, reducing the number of evaluated hyperparameter settings should be a general goal of the community.

The \textbf{main contribution} of this work is the introduction of a new research framework: \emph{Hyperparameter transfer across adjustments (HT-AA)}, which empowers developers with a third option:
\begin{itemize}
  \item[3.] Automatically transfer knowledge from previous HPOs.
\end{itemize}
This option leads to advantages in two aspects:
The automation of decision making and the utilization of previous knowledge.
On the one hand, the automation allows to benchmark strategies, replaces expensive manual decision making, and enables reproducible and comparable experiments;
on the other hand, utilizing previous knowledge leads to faster development cycles, lower costs, and reduced environmental impacts.

To lay a solid foundation for the new HT-AA framework, our \textbf{individual contributions} are as follows:
\begin{itemize}
    \item We formally introduce a basic version of the HT-AA problem (\Secref{sec:problem_statement}).
    \item We provide four simple baseline algorithms\footnote{Python package baselines: \href{https://github.com/hp-transfer/ht_optimizers/tree/v0.1.0}{\nolinkurl{github.com/hp-transfer/ht_optimizers/tree/v0.1.0}}} for our basic HT-AA problem (\Secref{sec:approach}).
    \item We provide a comprehensive set of eight novel benchmarks\footnote{Python package benchmarks: \href{https://github.com/hp-transfer/ht_benchmarks/tree/v0.1.0}{\nolinkurl{github.com/hp-transfer/ht_benchmarks/tree/v0.1.0}}} for our basic HT-AA problem (\Secref{sec:benchmarks}).
    \item We perform an empirical study on this set of benchmarks\footnote{Source code experiments: \href{https://github.com/hp-transfer/htaa_experiments/tree/v0.1.0}{\nolinkurl{github.com/hp-transfer/htaa_experiments/tree/v0.1.0}}}, showing that our simple baseline algorithms outperform HPO from scratch up to 1.2--2.6x on average depending on the budgets (\Secref{sec:experiments}).
    \item We relate the HT-AA framework to existing research efforts and discuss the research opportunities it opens up (\Secref{sec:related_work}).
    \item To facilitate future research on HT-AA, we provide open-source code for our experiments and benchmarks and provide a python package with an out-of-the-box usable implementation of our HT-AA algorithms.
\end{itemize}

\section{Hyperparameter Transfer Across Adjustments}\label{sec:problem_statement}

After presenting a broad introduction to the topic, we now provide a detailed description of hyperparameter transfer across developer adjustments (HT-AA).
We first introduce hyperparameter optimization, then discuss the types of developer adjustments, and finally describe the transfer across these adjustments.

\paragraph{Hyperparameter optimization (HPO)}
The HPO formulation we utilize in this work is as follows:
\begin{equation}
  \minimize_{\vx \in \mathcal{X}} f_{\mathcal{A}}(\vx) \quad \text{with $b$ evaluations} \qquad ,
\end{equation}
where $f_{\mathcal{A}}(\vx)$ is the objective function for ML algorithm $\mathcal{A}$ with hyperparameter setting $\vx$, $b$ is the number of available evaluations, and $\mathcal{X}$ is the search space.
We allow the search space $\mathcal{X}$ to contain categorical and numerical dimensions alike and consider only sequential evaluations.
We refer to a specific HPO problem with the 3-tuple $(\mathcal{X}, \,  f_{\mathcal{A}}, \, b)$.
For a discussion on potential extensions of our framework to different HPO formulations, we refer the reader to \Secref{sec:related_work}.

\paragraph{Developer adjustments}

\begin{figure}[t]
    \centering
    \includegraphics{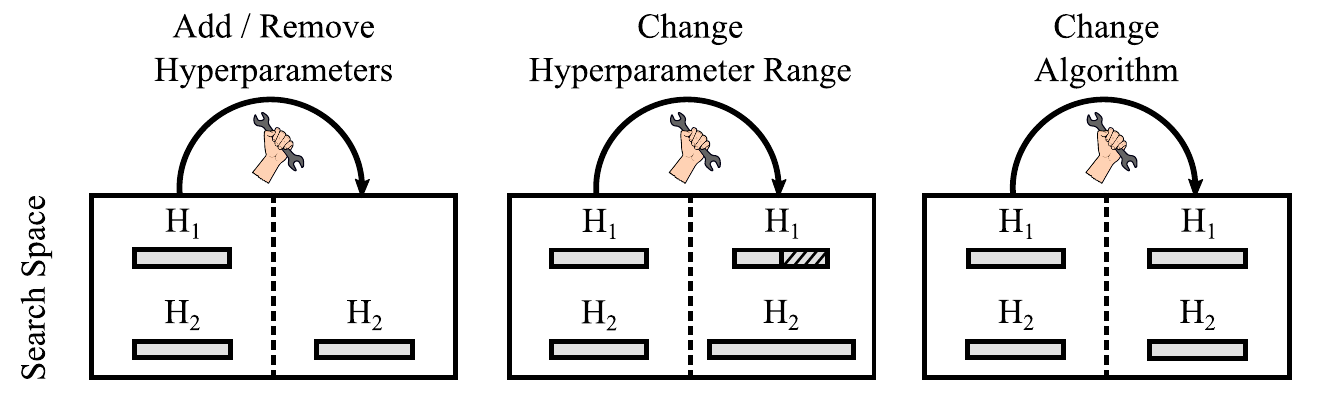}
    \caption{
        Developer adjustments from the perspective of hyperparameter optimization.
    }
\label{fig:adjustments}
\end{figure}

We now put developer adjustments on concrete terms and introduce a taxonomy of developer adjustments.
We consider two main categories of developer adjustments: ones that do not change the search space $\mathcal{X}$ (homogeneous adjustments) and ones that do (heterogenous adjustments).
Homogeneous adjustments could either change the algorithm's implementation or the hardware that the algorithm is run on.
Heterogeneous adjustments can be further categorized into adjustments that add or remove a hyperparameter (hyperparameter adjustments) and adjustments that change the search space for a specific hyperparameter (range adjustments).
\Figref{fig:adjustments} shows an illustration of the adjustment types.

\paragraph{Knowledge transfer across adjustments}
In general, a continuous stream of developer adjustments could be accompanied by multiple HPOs.
We simplify the problem in this fundamental work and only consider the transfer between two HPO problems; we discuss a potential extension in \Secref{sec:related_work}.
The two HPO problems arise from adjustments $\Psi$ to a ML algorithm $\mathcal{A}_{\text{old}}$ and its search space $\mathcal{X}_{\text{old}}$, which lead to $\mathcal{A}_{\text{new}}, \mathcal{X}_{\text{new}} := \Psi(\mathcal{A}_{\text{old}}, \, \mathcal{X}_{\text{old}})$.
Specifically, the hyperparameter transfer across adjustments problem is to solve the HPO problem $(\mathcal{X}_{\text{new}}, \, f_{\mathcal{A}_{\text{new}}}, \, b_{\text{new}})$, given the results for $(\mathcal{X}_{\text{old}}, \, f_{\mathcal{A}_{\text{old}}}, \, b_{\text{old}})$.
Compared to HPO from scratch, developers can choose a lower budget $b_{new}$, given evidence for a transfer algorithm achieving the same performance faster.

\section{Baseline Algorithms for HT-AA}\label{sec:approach}

In this section we present four baselines for the specific instantiation of the hyperparameter transfer across adjustments (HT-AA) framework discussed in \Secref{sec:problem_statement}.
We resist the temptation to introduce complex approaches alongside a new research framework and instead focus on a solid foundation.
Specifically, we focus on approaches that do not use any knowledge from the new HPO for the transfer.
We first introduce the basic HPO algorithm that the transfer approaches build upon then introduce notation for two decompositions of HPO search spaces across adjustments, and finally, we present the four baselines themselves.

\subsection{Preliminaries}

\paragraph{Background} For basic hyperparameter optimization and parts of the transfer algorithms, we employ the Tree-Structured Parzen Estimator (TPE) algorithm \citep{bergstra2011algorithms}, which is the default algorithm in the popular HyperOpt package \citep{bergstra2013making}.
TPE uses kernel density estimators to model the densities $l(\vx)$ and $g(\vx)$, for the probability of a given hyperparameter configuration $\vx$ being worse ($l$), or better ($g$), than the best already evaluated configuration.
To decide which configuration to evaluate, TPE then solves $\vx^* \in \argmax_{\vx \in \mathcal{X}}  \nicefrac{g(\vx)}{b(\vx)}$ approximately.
In our experiments, we use the TPE implementation and hyperparameter settings from \citet{falkner-icml2018}.

\paragraph{Search space decomposition: Hyperparameter adjustments}
For hyperparameter adjustments the new search space $\mathcal{X}_{\text{new}}$ and the old search space $\mathcal{X}_{\text{old}}$ only differ in hyperparameters, not in hyperparameter ranges, so we can decompose the search spaces as
$\mathcal{X}_{\text{new}} = \mathcal{X}_{\text{only-new}} \, \times \, \mathcal{X_{\text{both}}}$
and
$\mathcal{X}_{\text{old}} = \mathcal{X}_{\text{both}} \, \times \, \mathcal{X_{\text{only-old}}}$, where $\mathcal{X}_{\text{both}}$ is the part of the search space that remains unchanged across adjustments (see \Figref{fig:approach} for reference).
All baselines use this decomposition and project the hyperparameter settings that were evaluated in the old HPO from $\mathcal{X}_{\text{old}}$ to $\mathcal{X}_{\text{both}}$.

\paragraph{Search space decomposition: Range adjustments}
A range adjustment can remove values from the hyperparameter range or add values.
For an adjustment of hyperparameter range $\mathcal{X}^{H_i}_{old}$ to $\mathcal{X}^{H_i}_{\text{new}}$ this can be expressed as $\mathcal{X}^{H_i}_{\text{new}}
= \mathcal{X}^{H_i}_{\text{both}} \cup \mathcal{X}^{H_i}_{\text{both,range-only-new}}$ with $\mathcal{X}^{H_i}_{\text{both}} = \mathcal{X}^{H_i}_{old} \setminus \mathcal{X}^{H_i}_{\text{both,range-only-old}}$.

\subsection{Only Optimize New Hyperparameters}
A natural strategy for HT-AA is to set hyperparameters in $\mathcal{X}_{\text{both}}$ to the best setting of the previous HPO and only optimize hyperparameters in $\mathcal{X}_{\text{only-new}}$ \citep{agostinelli2014learning,huang2017densely,wu2018group}.
If the previous best setting is not a valid configuration anymore, i.e., it has values in $\mathcal{X}^{H_i}_{\text{both,range-only-old}}$ for a hyperparameter $H_i$ still in $\mathcal{X}_{\text{both}}$, this strategy uses the best setting that still is a valid configuration.
In the following, we refer to this strategy as \emph{only-optimize-new}.

\subsection{Drop Unimportant Hyperparameters}
A strategy inspired by manual HT-AA efforts is to only optimize important hyperparameters.
The utilization of importance statistics was, for example, explicitly disclosed in the seminal work on AlphaGo \citep{chen2018bayesian}.
Here, we determine the importance of each individual hyperparameter with functional analysis of variance (fANOVA) \citep{hutter-icml14a} and do not tune hyperparameters with below mean importance.
Therefore, this strategy only optimizes hyperparameters in $\mathcal{X}_{\text{only-new}}$ and hyperparameters in $\mathcal{X}_{\text{both}}$ with above mean importance.
In the following, we refer to this strategy as \emph{drop-unimportant}.

\subsection{First Evaluate Best}

The \emph{best-first} strategy uses only-optimize-new for the first evaluation, and uses standard TPE for the remaining evaluations.
This strategy has a large potential speedup and low risk as it falls back to standard TPE.

\subsection{Transfer TPE (T2PE)}\label{subsec:t2pe}

We introduce T2PE in two parts: first, the strategy to deal with homogeneous adjustments (unchanged search space) or hyperparameter adjustments (add/remove hyperparameters), and second, the strategy to deal with range adjustments.
Please find the pseudocode for T2PE in \Apxref{apx:pseudocode}.

\begin{figure}[t]
    \centering
    \includegraphics{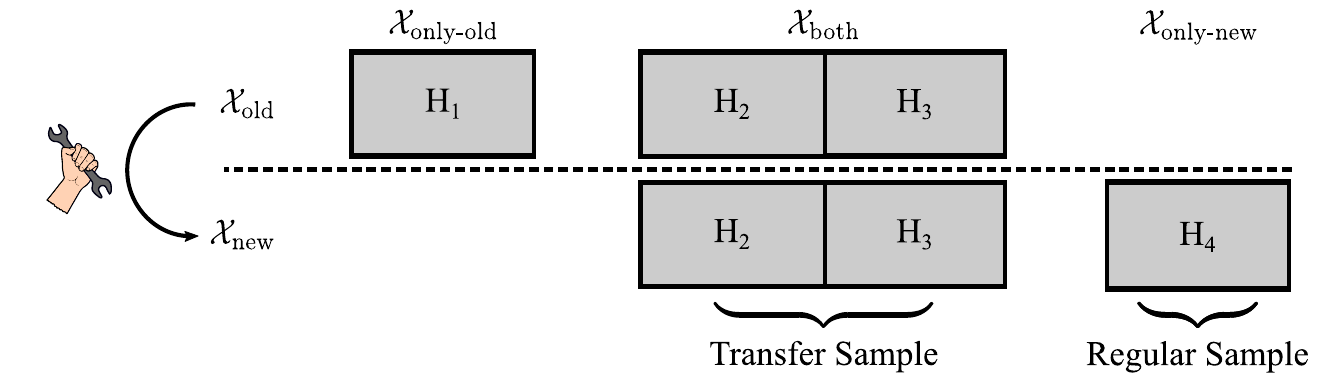}
    \caption{
        Example Search space decomposition for a hyperparameter addition and removal.
    }
\label{fig:approach}
\end{figure}

\paragraph{Homogeneous and hyperparameter adjustments}
Over $\mathcal{X}_{\text{both}}$ we sample from a TPE model fitted on the projected results of the previous HPO, and for $\mathcal{X}_{\text{only-new}}$ we use a random sample (\Figref{fig:approach}).
Once there are enough evaluations to fit a TPE model for the new HPO, we fit and use this new TPE model.
This is the case after $2(\dim(\mathcal{X}_{\text{new}}) + 1)$ evaluations for the TPE implementation we use.

\paragraph{Range adjustments}
We handle range removals ($\mathcal{X}^{H_i}_{\text{both,range-only-old}} \neq \emptyset$) separately from range addition ($\mathcal{X}^{H_i}_{\text{both,range-only-new}} \neq \emptyset$).
To handle range removals, T2PE ignores hyperparameter settings from the old HPO that have hyperparameter values in $\mathcal{X}_{\text{both,range-only-old}}$ when forming the model $M_{\text{both}}$.
The main idea in how we handle additions to ranges, is to guarantee that each added range $\mathcal{X}^{H_i}_{\text{both,range-only-new}}$ is sampled with probability proportional to its size with respect to $|\mathcal{X}^{H_i}_{\text{new}}|$, i.e., with probability $p_i = \frac{|\mathcal{X}^{H_i}_{\text{both,range-only-new}}|}{|\mathcal{X}^{H_i}_{\text{new}}|}$.
If there are log-uniform priors on the hyperparameter range, we take this prior into account when computing $p_i$.
To guarantee the above property, T2PE first samples $\rvx_{\text{both}}$ from $\mathcal{X_{\text{both}}}$ according to $M_{\text{both}}$, then mutates $\rvx^i_{\text{both}}$ with probability $p_i$ to a random sample from $\mathcal{X}^{H_i}_{\text{both,range-only-new}}$.

\section{Benchmarks for HT-AA}\label{sec:benchmarks}

We introduce eight novel benchmarks for the basic hyperparameter transfer across adjustments (HT-AA) problem discussed in \Secref{sec:problem_statement}.
As is common in hyperparameter optimization research, we employ tabular and surrogate benchmarks to allow cheap and reproducible benchmarking \citep{perrone2018scalable,falkner-icml2018}.
Tabular benchmarks achieve this with a lookup table for all possible hyperparameter settings.
In contrast, surrogate benchmarks fit a model for objective function \citep{eggensperger-metasel14a}.
We base our benchmarks on four existing hyperparameter optimization (HPO) benchmarks \citep{perrone2018scalable,klein2019tabular,dong2019bench},
which cover four different machine learning algorithms: a fully connected neural network (FCN), neural architecture search for a convolutional neural network (NAS), a support vector machine (SVM), and XGBoost (XGB).
For each of these base benchmarks, we consider two different types of adjustments (\Tabref{tbl:benchmark_adjustments}) to arrive at a total of eight benchmarks.
Additionally, for each algorithm and adjustment, we consider multiple tasks in our benchmarks.
Further, we provide a python package with all our benchmarks and refer the reader to \Apxref{apx:benchmark_suite} for additional details on the benchmarks.

\begin{table}[ht]
  \caption{Developer adjustments in benchmarks}
  \centering
  \begin{tabularx}{0.95\textwidth}{lX}
    \toprule
    Benchmark & Adjustments \\
    \midrule
    FCN-A & Increase \#units-per-layer 16$\times$; Double \#epochs;
    Fix batch size hyperparameter\\
    FCN-B & Add per-layer choice of activation function; Change learning rate schedule \\ \midrule
    NAS-A & Add 3x3 average pooling as choice of operation to each edge \\
    NAS-B & Add node to cell template (adds 3 hyperparameters) \\ \midrule
    XGB-A & Expose four booster hyperparameters \\
    XGB-B & Change four unexposed booster hyperparameter values \\ \midrule
    SVM-A & Change kernel; Remove hyperparameter for old kernel;
    \newline Add hyperparameter for new kernel \\
    SVM-B & Increase range for cost hyperparameter \\
    \bottomrule
  \end{tabularx}
  \label{tbl:benchmark_adjustments}
\end{table}

\section{Experiments and Results}\label{sec:experiments}

In this section, we empirically evaluate the four baseline algorithms presented in \Secref{sec:approach} as solutions for the hyperparameter transfer across adjustments problem.
We first describe the evaluation protocol used through all studies and then present the results.

\paragraph{Evaluation protocol}
We use the benchmarks introduced in \Secref{sec:benchmarks} and focus on the speedup of transfer strategies over TPE.
Specifically, we measured how much faster a transfer algorithm reaches a given objective value compared to TPE in terms of the number of evaluations.
We repeated all measurements across 100 different random seeds and report results for validation objectives, as not all benchmarks provide test objectives, and to reduce noise in our evaluation.
We terminate runs after 400 evaluations and report ratio of means.
To aggregate these ratios across tasks and benchmarks, we use the geometric mean.
To determine the target objective values, we measured TPE's average performance for 10, 20, and 40 evaluations.
We chose this range of evaluations as a survey among NeurIPS2019 and ICLR2020 authors indicates that most hyperparameter optimizations (HPOs) do not consider more than 50 evaluations \citep{bouthillier2020Survey}.
Further, for transfer approaches, we perform this experiment for different evaluation budgets for the HPO before the adjustments (also for 10, 20, and 40 evaluations).

\begin{figure}[t]
    \centering
    \includegraphics{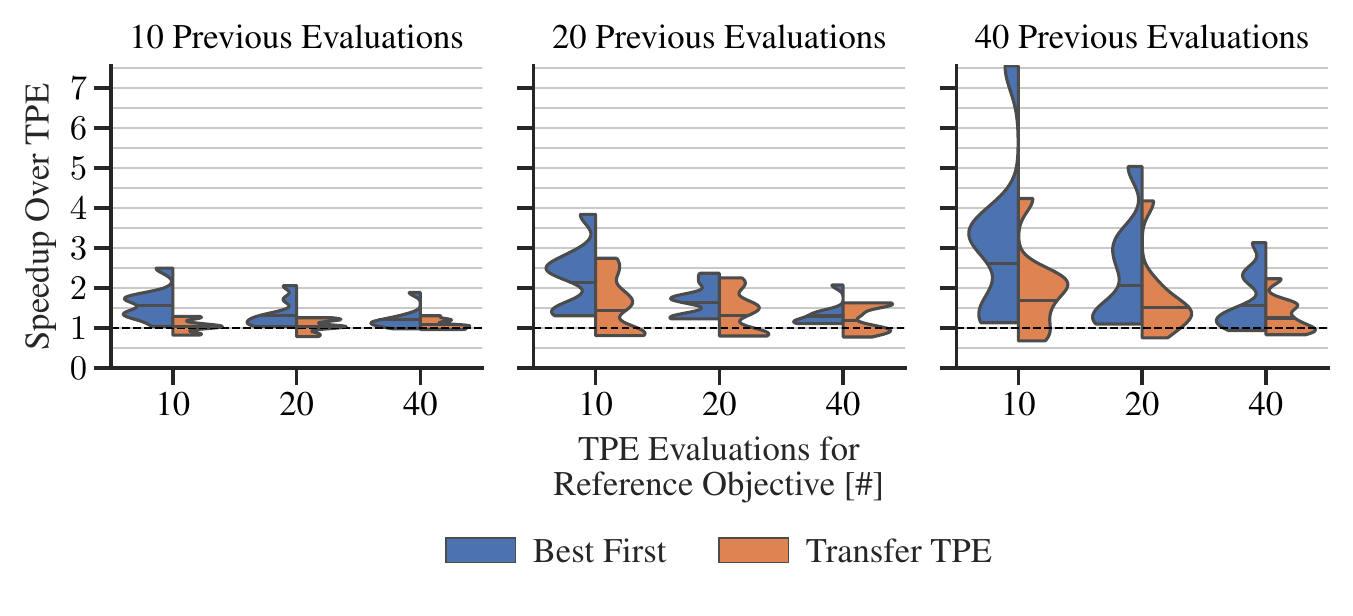}
    \vspace{-10pt}
    \caption{Speedup to reach a given reference objective value compared to TPE for best-first and transfer TPE across 8 benchmarks. The violins estimate densities of benchmark geometric means. The horizontal line in each violin shows the geometric mean across these benchmark means. \#Evaluations for the old HPO increases from left to right. The x-axis shows the budget for the TPE reference.}
    \label{fig:global_speedup_transfer_tpe_no_best_first+transfer_tpe_no_ttpe_over_tpe}
\end{figure}

\begin{figure}[t]
    \centering
    \includegraphics{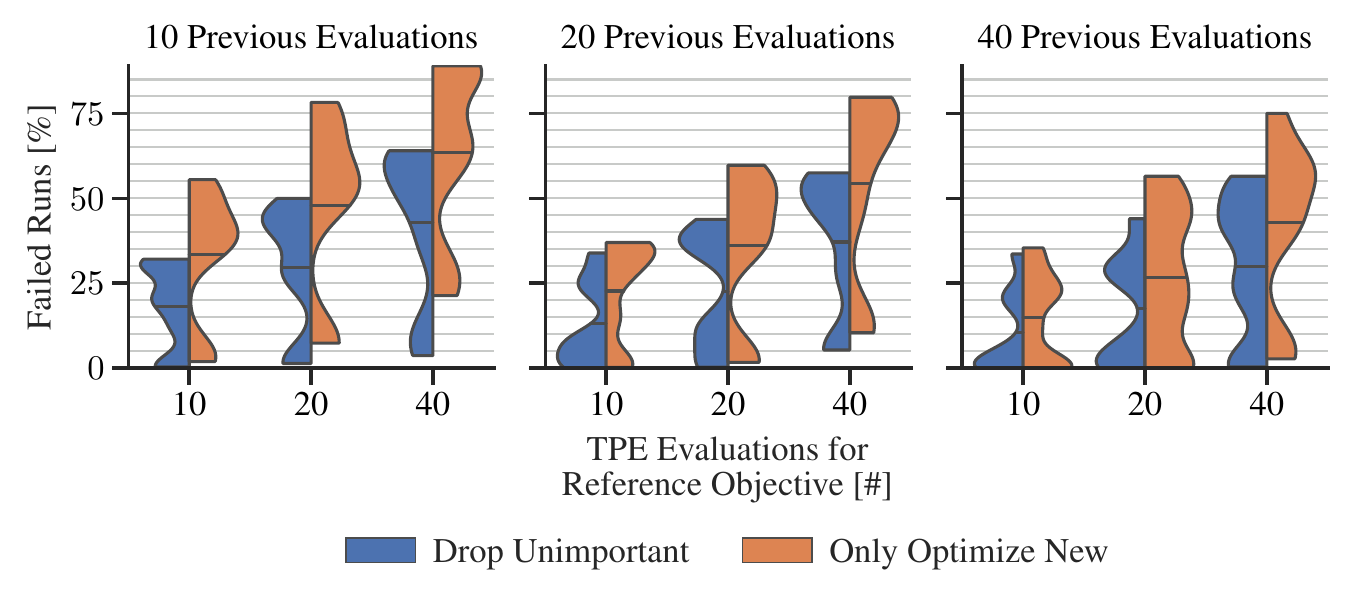}
    \vspace{-10pt}
    \caption{Percent of runs that do not reach the reference objective for drop-unimporant and only-optimize-new.
    Each data point for the violins represents the mean percentage of failures for a benchmark.
    The line in each violin shows the mean across these benchmark means.
    \#Evaluations for the old HPO increases from left to right.
    The x-axis shows the budget for the TPE reference.}
    \label{fig:global_nan_percent_transfer_top_and_transfer_importance}
\end{figure}

\paragraph{Results}
The transfer TPE (T2PE) and best-first strategy lead to large speedups, while drop-unimportant and only-optimize-new perform poorly.
On average and depending on the budgets for the old and new HPO, T2PE reaches the given objective values 1.0--1.7x faster than TPE, and best-first 1.2--2.6x faster
(\Figref{fig:global_speedup_transfer_tpe_no_best_first+transfer_tpe_no_ttpe_over_tpe}, \Tabref{tbl:speedups}).
As T2PE  and best-first work well on their own, a natural idea is to combine them.
The combination leads to further speedups over best-first if the budget for the old HPO was 20 or 40 (on average about 0.1 more speedup; \Apxref{apx:combined_speedup}, \Tabref{tbl:speedups}).
There are two main trends visible: (1) The more optimal the target objective, the smaller the speedup, and (2) the higher the budget for the previous HPO, the higher the speedup.
For a more fine-grained visualization that shows violin plots over task means for each benchmark, we refer to \Apxref{apx:detailed_main}.
Drop-unimportant and only-optimize-new do not reach the performance of TPE in a large percentage of cases, even while given 10x the budget compared to TPE
(\Figref{fig:global_nan_percent_transfer_top_and_transfer_importance}).
These high failure rates make an evaluation for the speedup unfeasible.
For the failure rates for TPE, T2PE, and best-first (0-6\%) we refer the reader to \Apxref{apx:main_failure}.

\begin{table}[t]
  \caption{Average speedup across benchmarks for different \#evaluations for the old and new HPO.}
  \centering
  \begin{tabular}{ccccc}
    \toprule
    \#Evals Old & \#Evals New & Best First & Transfer TPE & Best First + Transfer TPE \\
    \midrule
    $10$ & $10$ & \textbf{1.6} & 1.0 & 1.5 \\
         & $20$ & \textbf{1.3} & 1.0 & \textbf{1.3} \\
         & $40$ & \textbf{1.2} & 1.1 & \textbf{1.2} \\  \midrule
    $20$ & $10$ & 2.1 & 1.4 & \textbf{2.3} \\
     & $20$ & 1.6 & 1.3 & \textbf{1.9} \\
     & $40$ & 1.3 & 1.2 & \textbf{1.4} \\ \midrule
     $40$ & $10$ & 2.6 & 1.7 & \textbf{2.9} \\
       & $20$ & 2.1 & 1.5 & \textbf{2.3} \\
       & $40$ & 1.6 & 1.3 & \textbf{1.7} \\
    \bottomrule
  \end{tabular}
  \label{tbl:speedups}
\end{table}

Additionally, we provide a study on the improvement in objective value for a fixed number of evaluations in \Apxref{apx:improvement_analysis}; in \Apxref{apx:control_study_tpe2} we show the results of a control study that compares TPE with different ranges of random seeds; and in \Apxref{apx:control_study_random} we compare random search to TPE.

\section{Related Work and Research Opportunities}\label{sec:related_work}

In this section, we discuss work related to hyperparameter transfer across adjustments (HT-AA) and present several research opportunities in combining existing ideas with HT-AA.

\paragraph{Transfer learning}
Transfer learning studies how to use observations from one or multiple source tasks to improve learning on one or multiple target tasks \citep{zhuang2019comprehensive}.
If we view the HPO problems before and after specific developer adjustments as tasks, we can consider HT-AA as a specific transfer learning problem.
As developer adjustments may change the search space, HT-AA would then be categorized as a heterogeneous transfer learning problem \citep{day2017survey}.

\paragraph{Transfer learning across adjustments}
Recently, \citet{openai2019dota} transferred knowledge between deep reinforcement learning agents across developer adjustments.
They crafted techniques to preserve, or approximately preserve, the neural network policy for each type of adjustment they encountered.
Their transfer strategies are inspired by Net2Net knowledge transfer \citep{chen2015net2net}, and they use the term surgery to refer to this practice.
Their work indicates that transfer learning across adjustments is not limited to knowledge about hyperparameters, but extends to a more general setting, leaving room for many research opportunities.

\paragraph{Continuous knowledge transfer}
In this paper, we focus on transferring knowledge from the last HPO performed, but future work could investigate a continuous transfer of knowledge across many cycles of adjustments and HPOs.
Transferring knowledge from HPO runs on multiple previous versions could lead to further performance gains, as information from each version could be useful for the current HPO.
Such \emph{continuous} HT-AA would then be related to the field of continual learning \citep{THRUN199525,delange2020continual}.

\paragraph{Hyperparameter transfer across tasks (HT-AT)}
There exists an extensive research field that studies the transfer across \emph{tasks} for HPOs \citep{vanschoren2018meta}.
The main difference to hyperparameter transfer across \emph{adjustments} is that the former assumes an unchanging search space, whereas dealing with such changes is one of the main challenges in HT-AA.
In HT-AT, the search space and the ML algorithm remain unchanged, but the task that the algorithm is applied to changes.
Another difference is that most work on HT-AT considers large amounts of meta-data; up to more than a thousand tasks and function evaluations \citep{NIPS2018_8248,metz2020using}.

\emph{Homogeneous} hyperparameter transfer across adjustments (homogeneous HT-AA) problems, where none of the adjustments changes the search space, are syntactically equivalent to HT-AT problems.
For this homogeneous HT-AA, existing approaches for HT-AT could, in principle, be applied without modification; this includes, for example the transfer acquisition function~\citep{wistuba2018scalable}, multi-task bayesian optimization~\citep{swersky2013multi}, multi-task adaptive bayesian linear regression ~\citep{perrone2018scalable}, ranking-weighted gaussian process ensemble~\citep{feurer2018scalable}, and difference-modelling bayesian optimisation~\citep{pmlr-v54-shilton17a}.

Further, an adaptation of across-task strategies to the across-adjustments setting could lead to more powerful HT-AA approaches in the future. Finally, the combination of across-task and across-adjustments hyperparameter transfer is an exciting research opportunity that could provide even larger speedups than either transfer strategy on its own.

\paragraph{Advanced hyperparameter optimization}
HT-AA can be combined with one of the many extensions to the basic hyperparameter optimization (HPO) formulation.
One such extension is multi-fidelity HPO, which allows the use of cheap-to-evaluate approximations to the actual objective \citep{li-iclr17,falkner-icml2018}.
Similarly, cost-aware HPO adds a cost to each hyperparameter setting, so a cost model can prioritize the evaluation of cheap hyperparameter settings over expensive ones \citep{snoek-nips12a}.
Yet another extension is to take different kinds of evaluation noise into account \citep{kersting2007most} or to consider not one, but multiple objectives to optimize for \citep{khan2002multi}.
All these HPO formulations can be studied in conjunction with HT-AA, to either provide further speedups or deal with more general optimization problems.

\paragraph{Guided machine learning}
The field of guided machine learning (gML) studies the design of interfaces that enables humans to guide ML processes \citep{Westphal2019case}.
An HT-AA algorithm could be viewed as a ML algorithm that receives incremental guidance in the form of arbitrary developer adjustments; the interface would then be the programming language(s) the ML algorithm is implemented in.

On a different note, gML could provide HT-AA algorithms with additional information about the adjustments to the ML algorithm.
For example, when adding a hyperparameter, there are two distinctions we can make: Either an existing hyperparameter is exposed
(e.g., the dropout rate was previously hardcoded as 0.5, and is now tuned)
, or a new component is added to the algorithm that introduces a new hyperparameter
(e.g., a new learning rate schedule that introduces a decay hyperparameter)
.
From the HPO problem itself, we cannot know which case it is, and neither which fixed value an exposed hyperparameter had.
Guided HT-AA algorithms could ask for user input to fill this knowledge gap. Alternatively, HT-AA algorithms with code analysis could automatically extract this knowledge from the source code.

\paragraph{Programming by optimization} Relatedly, the programming by optimization (PbO) framework~\citep{hoos2012programming} proposes the automatic construction of a search space of algorithms, based on code annotations, and the subsequent automated search in this search space.
While this framework considers evolving search spaces over incremental developer actions, each task and development step restarts the search from scratch.
This is in contrast to our hyperparameter transfer framework that alleviates the need to restart from scratch after each developer adjustment.

\section{Conclusion}

In this work, we introduced hyperparameter transfer across developer adjustments to improve efficiency during the development of ML algorithms.
In light of rising energy demands of ML algorithms and rising global temperatures, more efficient ML development practices are an important issue now and will become more important in the future.
As already two of the simple baseline algorithm considered in this work lead to large empirical speedups, our new framework represents a promising step towards efficient ML development.

\section*{Acknowledgements}

The authors acknowledge support by the state of Baden-W\"{u}rttemberg through bwHPC and the German Research Foundation (DFG) through grant no INST 39/963-1 FUGG.
Robert Bosch GmbH is acknowledged for financial support. 
This work has partly been supported by the European Research Council (ERC) under the European Union’s Horizon 2020 research and innovation programme under grant no.\ 716721.

\bibliography{bibliography/lib,bibliography/local,bibliography/proc}
\bibliographystyle{utils/iclr2021_conference}

\newpage
\appendix

\section{Pseudocode}\label{apx:pseudocode}

\begin{algorithm}[H]
\caption{Sampling strategy in transfer TPE}
\hspace*{\algorithmicindent} \textbf{Input}: Current search space $\mathcal{X}_{\text{new}}$, previous search space $\mathcal{X}_{\text{old}}$, \newline
\hspace*{1.6cm} config ranking of previous optimization $\mathcal{C}$, prior over $\mathcal{X}_{\text{new}}$
\begin{algorithmic}[1]
    \State Decompose $\mathcal{X}_{\text{new}} = ( \mathcal{X}_{\text{both}} \, \cup \, \mathcal{X}_{\text{both,range-only-new}}) \, \times \, \mathcal{X_{\text{only-new}}} $
    \State Discard configs in $\mathcal{C}$ that have hyperparameter values in $\mathcal{X}_{\text{both,range-only-new}}$
    \State Project configs in $\mathcal{C}$ to space $\mathcal{X_{\text{both}}}$, to yield config ranking $\mathcal{C}_{\text{both}}$
    \State Fit TPE model $M_{\text{both}}$ for $\mathcal{X_{\text{both}}}$ on $\mathcal{C}_{\text{both}}$
    \For{$t$ \textbf{in}  $1, \, \ldots, \, N$}
      \If{is random fraction}  \Comment{From TPE implementation, e.g., 1/3 of cases}
          \State Sample $\rvx_{\text{new}}$ from prior on $\mathcal{X}_{\text{new}}$
      \ElsIf{no model for $\mathcal{X}_{\text{new}}$}
          \State Sample $\rvx_{\text{both}}$ from $\mathcal{X_{\text{both}}}$ according to $M_{\text{both}}$
          \For{hyperparameter range $\mathcal{X}^{H_i}_{\text{both,range-only-new}} \neq \emptyset$ \textbf{in} $\mathcal{X}_{\text{both,range-only-new}}$}
              \State Set $p := \frac{
                |\mathcal{X}^{H_i}_{\text{both,range-only-new}}|
                }{
                |\mathcal{X}^{H_i}_{\text{new}}|
                }$  \Comment{Take into account priors}
               \State Sample $\rx^i$ from prior on $\mathcal{X}^{H_i}_{\text{both,range-only-new}}$
               \State Set $\rvx^i_{\text{both}} := \rx^i$ with probability $p$
          \EndFor

          \State Sample $\rvx_{\text{only-new}}$ from prior on $\mathcal{X_{\text{only-new}}}$
          \State Combine $\rvx_{\text{both}}$ with $\rvx_{\text{only-new}}$ to yield sample $\rvx_{\text{new}}$
      \Else
          \State Fit TPE model $M_{\text{new}}$ for $\mathcal{X_{\text{new}}}$ on current observations
          \State Sample $\rvx_{\text{new}}$ from $\mathcal{X}_{\text{new}}$ according to $M_{\text{new}}$
      \EndIf
    \EndFor
    \Return
\end{algorithmic}
\label{alg:pseudocode}
\end{algorithm}

\section{Benchmark Suite Details}\label{apx:benchmark_suite}

\subsection{Overview}

\begin{table}[H]
  \caption{Benchmarks overview}
  \centering
  \begin{tabular}{lrrr}
    \toprule
    Benchmark &  \#Hyperparameters Old & \#Hyperparameters New & \#Tasks \\
    \midrule
    FCN-A & $6$ & $5$ & $4$ \\
    FCN-B & $6$ & $8$ & $4$ \\ \midrule
    NAS-A & $6$ & $6$ & $3$ \\
    NAS-B & $3$ & $6$ & $3$ \\ \midrule
    XGB-A & $5$ & $9$ & $10$ \\
    XGB-B & $6$ & $6$ & $10$ \\  \midrule
    SVM-A & $2$ & $2$ & $10$ \\
    SVM-B & $2$ & $2$ & $10$ \\
    \bottomrule
  \end{tabular}
  \label{tbl:benchmark_overview}
\end{table}

\subsection{FCN-A \& FCN-B}

\paragraph{Budget} For FCN-A the budget is set to 100.
For FCN-B, additional to the changes in the search space (\Tabref{tbl:fcnb}), the budget is increased from 50 to 100 epochs.

\begin{table}[H]
  \caption{Values for integer coded hyperparameters in FCN benchmarks}
  \centering
  \begin{tabular}{lc}
    \toprule
    Hyperparameter & Values \\
    \midrule
    \# Units Layer \{1,\,2\}   &    $( 16, \, 32, \, 64, \, 128, \, 256, \, 512 )$  \\
    Dropout Layer \{1,\,2\}    &    $( 0.0, \, 0.3, \, 0.6 )$   \\
    Initial Learning Rate    &    $( 0.0005, \, 0.001, \, 0.005, \, 0.01, \, 0.05, \, 0.1 )$ \\
    Batch Size    &   $( 8, \, 16, \, 32, \, 64 )$ \\
    \bottomrule
  \end{tabular}
\label{tbl:fcn_detail}
\end{table}

\begin{table}[H]
  \caption{Search spaces in FCN-A. Numerical hyperparameters are encoded as integers, see \Tabref{tbl:fcn_detail} for specific values for these hyperparameters.}
  \centering
  \begin{tabular}{llcc}
    \toprule
    Steps & Hyperparameter & Range/Value & Prior \\
    \midrule
    $1$ & \# Units Layer 1    &    1    &   - \\
    $1$ & \# Units Layer 2    &    1    &   - \\
    $1$ & Batch Size    &    $\{ 0, \, \ldots, \, 3 \}$    &   Uniform \\
    \midrule
    $1$, $2$ & Dropout Layer  1   &    $\{ 0, \, \ldots, \, 2 \}$    &   Uniform \\
    $1$, $2$ & Dropout Layer  2   &    $\{ 0, \, \ldots, \, 2 \}$    &   Uniform \\
    $1$, $2$ & Activation Layer 1    &    $\{ \text{ReLu, tanh} \}$    &   Uniform \\
    $1$, $2$ & Activation Layer 2    &    $\{ \text{ReLu, tanh} \}$    &   Uniform \\
    $1$, $2$ & Initial Learning Rate    &    $\{ 0, \, \ldots, \, 5 \}$    &   Uniform \\
    $1$, $2$ & Learning Rate Schedule    &   Constant    &   Uniform \\
    \midrule
    $2$ & \# Units Layer 1    &    5    &   - \\
    $2$ & \# Units Layer 2    &    5    &   - \\
    $2$ & Batch Size    &    1    &   - \\

    \bottomrule
  \end{tabular}
\label{tbl:fcna}
\end{table}

\begin{table}[H]
  \caption{Search spaces in FCN-B. Numerical hyperparameters are encoded as integers, see \Tabref{tbl:fcn_detail} for specific values for these hyperparameters.}
  \centering
  \begin{tabular}{llcc}
    \toprule
    Steps & Hyperparameter & Range/Value & Prior \\
    \midrule
    $1$ & Activation Layer 1    &    tanh   &   - \\
    $1$ & Activation Layer 2    &    tanh   &   - \\
    $1$ & Learning Rate Schedule    &  Constant    &   - \\
    \midrule
    $1$, $2$ & \# Units Layer 1    &    $\{ 0, \, \ldots, \, 5 \}$    &   Uniform \\
    $1$, $2$ & \# Units Layer 2    &    $\{ 0, \, \ldots, \, 5 \}$    &   Uniform \\
    $1$, $2$ & Dropout Layer  1   &    $\{ 0, \, \ldots, \, 2 \}$    &   Uniform \\
    $1$, $2$ & Dropout Layer  2   &    $\{ 0, \, \ldots, \, 2 \}$    &   Uniform \\
    $1$, $2$ & Initial Learning Rate    &    $\{ 0, \, \ldots, \, 5 \}$    &   Uniform \\
    $1$, $2$ & Batch Size    &    $\{ 0, \, \ldots, \, 3 \}$    &   Uniform \\
    \midrule
    $2$ & Activation Layer 1    &    $\{ \text{ReLu, tanh} \}$    &   Uniform \\
    $2$ & Activation Layer 2    &    $\{ \text{ReLu, tanh} \}$    &   Uniform \\
    $2$ & Learning Rate Schedule    &  Cosine    &   - \\
    \bottomrule
  \end{tabular}
\label{tbl:fcnb}
\end{table}

\subsection{NAS-A \& NAS-B}

\begin{table}[H]
  \caption{Search spaces in NAS-A.}
  \centering
  \begin{tabular}{llcc}
    \toprule
    Steps & Hyperparameter & Range/Value & Prior \\
    \midrule
    $1$, $2$   &   $0\rightarrow2$ & \{ \text{none}, \text{skip-connect}, \text{conv1x1}, \text{conv3x3}, \text{avg-pool3x3} \} & Uniform \\
    $1$, $2$   &   $0\rightarrow3$ & \{ \text{none}, \text{skip-connect}, \text{conv1x1}, \text{conv3x3}, \text{avg-pool3x3} \} & Uniform \\
    $1$, $2$   &   $2\rightarrow3$ & \{ \text{none}, \text{skip-connect}, \text{conv1x1}, \text{conv3x3}, \text{avg-pool3x3} \} & Uniform \\ \midrule
    $2$   &   $0\rightarrow1$ & \{ \text{none}, \text{skip-connect}, \text{conv1x1}, \text{conv3x3}, \text{avg-pool3x3} \} & Uniform \\
    $2$   &   $1\rightarrow2$ & \{ \text{none}, \text{skip-connect}, \text{conv1x1}, \text{conv3x3}, \text{avg-pool3x3} \} & Uniform \\
    $2$   &   $1\rightarrow3$ & \{ \text{none}, \text{skip-connect}, \text{conv1x1}, \text{conv3x3}, \text{avg-pool3x3} \} & Uniform \\
    \bottomrule
  \end{tabular}
\label{tbl:nasa}
\end{table}

\begin{table}[H]
  \caption{Search spaces in NAS-B.}
  \centering
  \begin{tabular}{llcc}
    \toprule
    Steps & Hyperparameter & Range/Value & Prior \\
    \midrule
    $1$   &   $0\rightarrow1$ & \{ \text{none}, \text{skip-connect}, \text{conv1x1}, \text{conv3x3} \} & Uniform \\
    $1$  &   $0\rightarrow2$ & \{ \text{none}, \text{skip-connect}, \text{conv1x1}, \text{conv3x3} \} & Uniform \\
    $1$  &   $0\rightarrow3$ & \{ \text{none}, \text{skip-connect}, \text{conv1x1}, \text{conv3x3} \} & Uniform \\
    $1$   &   $1\rightarrow2$ & \{ \text{none}, \text{skip-connect}, \text{conv1x1}, \text{conv3x3} \} & Uniform \\
    $1$   &   $1\rightarrow3$ & \{ \text{none}, \text{skip-connect}, \text{conv1x1}, \text{conv3x3} \} & Uniform \\
    $1$  &    $2\rightarrow3$ & \{ \text{none}, \text{skip-connect}, \text{conv1x1}, \text{conv3x3} \} & Uniform
    \\\midrule
    $2$   &   $0\rightarrow1$ & \{ \text{none}, \text{skip-connect}, \text{conv1x1}, \text{conv3x3}, \text{avg-pool3x3} \} & Uniform \\
    $2$  &   $0\rightarrow2$ & \{ \text{none}, \text{skip-connect}, \text{conv1x1}, \text{conv3x3}, \text{avg-pool3x3} \} & Uniform \\
    $2$  &   $0\rightarrow3$ & \{ \text{none}, \text{skip-connect}, \text{conv1x1}, \text{conv3x3}, \text{avg-pool3x3} \} & Uniform \\
    $2$   &   $1\rightarrow2$ & \{ \text{none}, \text{skip-connect}, \text{conv1x1}, \text{conv3x3}, \text{avg-pool3x3} \} & Uniform \\
    $2$   &   $1\rightarrow3$ & \{ \text{none}, \text{skip-connect}, \text{conv1x1}, \text{conv3x3}, \text{avg-pool3x3} \} & Uniform \\
    $2$  &    $2\rightarrow3$ & \{ \text{none}, \text{skip-connect}, \text{conv1x1}, \text{conv3x3}, \text{avg-pool3x3} \} & Uniform \\
    \bottomrule
  \end{tabular}
\label{tbl:nasb}
\end{table}

\subsection{SVM-A \& SVM-B}

\begin{table}[H]
  \caption{Search spaces in SVM-A.}
  \centering
  \begin{tabular}{llcc}
    \toprule
    Steps & Hyperparameter & Range/Value & Prior \\
    \midrule
    $1$ & Kernel    &    Radial   &   - \\
    $1$ & Degree    &  $\{ 2, \, \ldots, \, 5 \}$     &   Uniform \\
    \midrule
    $1$, $2$ & Cost    &    $[2^{-10}, \, 2^{10}]$   &   Log-uniform \\
    \midrule
    $2$ & Kernel    &    Polynomial   &   - \\
    $2$ & $\gamma$    &    $[2^{-5}, \, 2^5]$   &   Log-uniform \\
    \bottomrule
  \end{tabular}
\label{tbl:svma}
\end{table}

\begin{table}[H]
  \caption{Search spaces in SVM-B.}
  \centering
  \begin{tabular}{llcc}
    \toprule
    Steps & Hyperparameter & Range/Value & Prior \\
    \midrule
    $1$ & Cost    &    $[2^{-5}, \, 2^{5}]$   &   Log-uniform \\
    \midrule
    $1$, $2$ & $\gamma$    &    1   &   - \\
    $1$, $2$ & Degree    &  5     &   - \\
    $1$, $2$ & Kernel    &    $\{ \text{Polynomial, Linear, Radial} \}$   &   Uniform \\
    \midrule
    $2$ & Cost    &    $[2^{-10}, \, 2^{10}]$   &   Log-uniform \\
    \bottomrule
  \end{tabular}
\label{tbl:svmb}
\end{table}

\subsection{XGB-A \& XGB-B}

\begin{table}[H]
  \caption{Search spaces in XGB-A}
  \centering
  \begin{tabular}{clcc}
    \toprule
    Steps & Hyperparameter & Range/Value & Prior \\
    \midrule
    $1$  &  Colsample-by-tree   &  1  &   - \\
    $1$  &  Colsample-by-level   &  1   &   - \\
    $1$  &  Minimum child weight   &   1   &   - \\
    $1$  &  Maximum depth      &   6   &   - \\
    \midrule
    $1$, $2$  &  Booster   &  Tree   &   - \\
    $1$, $2$  &  \# Rounds      &   $\{1, \, \ldots, \, 5,000 \}$   &   Uniform \\
    $1$, $2$  &  Subsample   &   $[0, \, 1]$   &   Uniform \\
    $1$, $2$  &  Eta   &   $[2^{-10}, \, 2^0]$   &   Log-uniform \\
    $1$, $2$  &  Lambda   &   $[2^{-10}, \, 2^{10}]$   &   Log-uniform \\
    $1$, $2$  &  Alpha   &   $[2^{-10}, \, 2^{10}]$   &   Log-uniform \\
    \midrule
    $2$  &  Colsample-by-tree   &   $[0, \, 1]$   &   Uniform \\
    $2$  &  Colsample-by-level   &   $[0, \, 1]$   &   Uniform \\
    $2$  &  Minimum child weight   &   $[2^0, \, 2^7]$   &   Log-uniform \\
    $2$  &  Maximum depth      &   $\{ 1, \, \ldots, \, 15 \}$   &   Uniform \\
    \bottomrule
  \end{tabular}
\label{tbl:xgba}
\end{table}

\begin{table}[H]
  \caption{Search spaces in XGB-B}
  \centering
  \begin{tabular}{clcc}
    \toprule
    Steps & Hyperparameter & Range/Value & Prior \\
    \midrule
    $1$  &  Colsample-by-tree   &  1  &   - \\
    $1$  &  Colsample-by-level   &  1   &   - \\
    $1$  &  Minimum child weight   &   1   &   - \\
    $1$  &  Maximum depth      &   6   &   - \\
    \midrule
    $1$, $2$  &  Booster   &  \{ Linear, Tree \}   &   - \\
    $1$, $2$  &  \# Rounds      &   $\{1, \, \ldots, \, 5,000 \}$   &   Uniform \\
    $1$, $2$  &  Subsample   &   $[0, \, 1]$   &   Uniform \\
    $1$, $2$  &  Eta   &   $[2^{-10}, \, 2^0]$   &   Log-uniform \\
    $1$, $2$  &  Lambda   &   $[2^{-10}, \, 2^{10}]$   &   Log-uniform \\
    $1$, $2$  &  Alpha   &   $[2^{-10}, \, 2^{10}]$   &   Log-uniform \\
    \midrule
    $2$  &  Colsample-by-tree   &  1  &   - \\
    $2$  &  Colsample-by-level   &  0.5   &   - \\
    $2$  &  Minimum child weight   &   10   &   - \\
    $2$  &  Maximum depth      &   10   &   - \\
    \bottomrule
  \end{tabular}
\label{tbl:xgbb}
\end{table}

\section{Detailed Speedups}\label{apx:detailed_main}

\begin{figure}[H]
    \centering
    \includegraphics{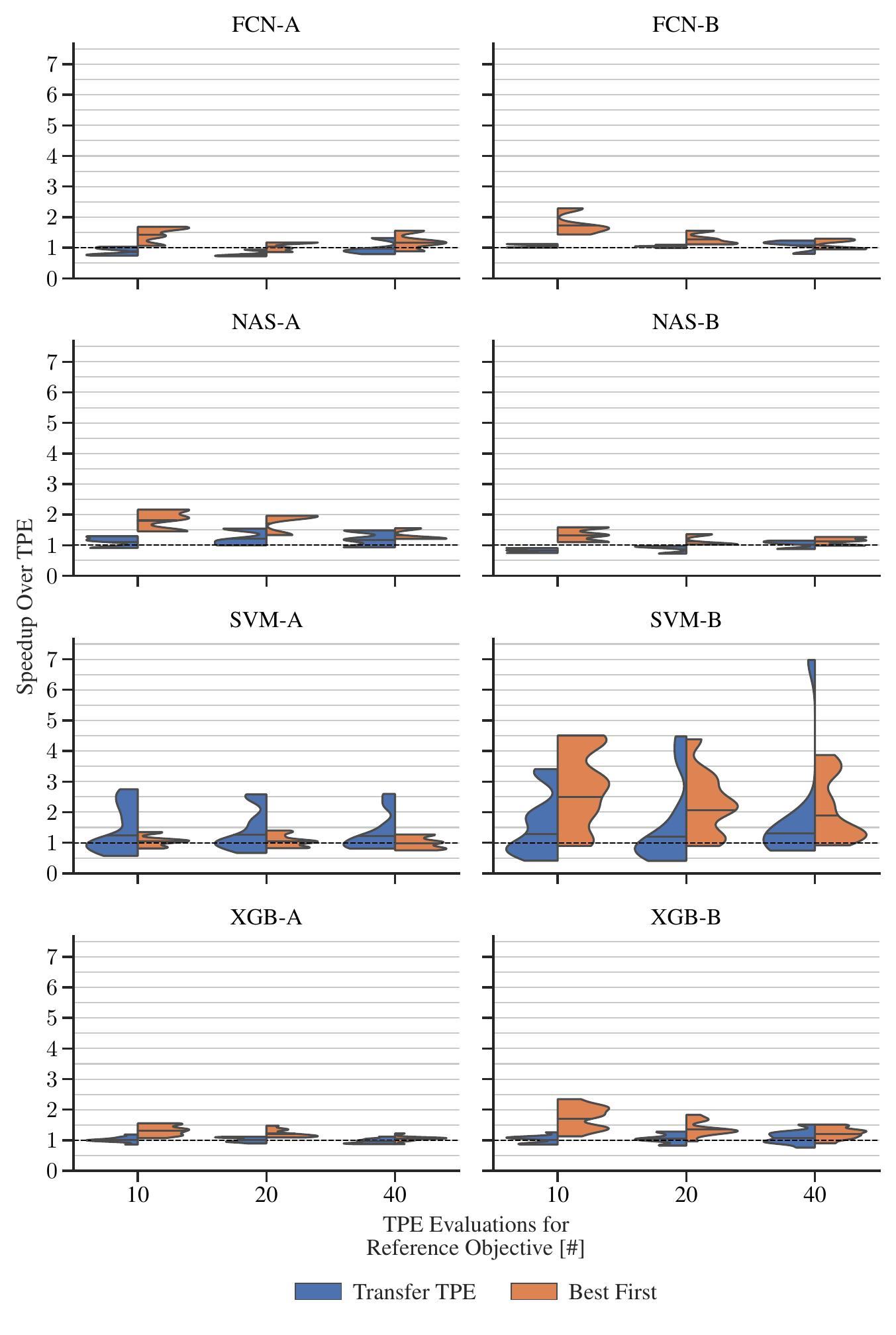}
    \vspace{-15pt}
    \caption{Speedup of transfer TPE and best-first over TPE across tasks for each of 8 benchmarks. The previous HPO has a budget of 10 evaluations here. The violins estimate densities of the task geometric means. The horizontal line in each violin shows the geometric mean across these task means. The x-axis shows the budget for the TPE reference.}
    \label{fig:speedup_transfer_tpe_no_best_first+transfer_tpe_no_ttpe_over_tpe_eval_ref_10}
\end{figure}

\begin{figure}[H]
    \centering
    \includegraphics{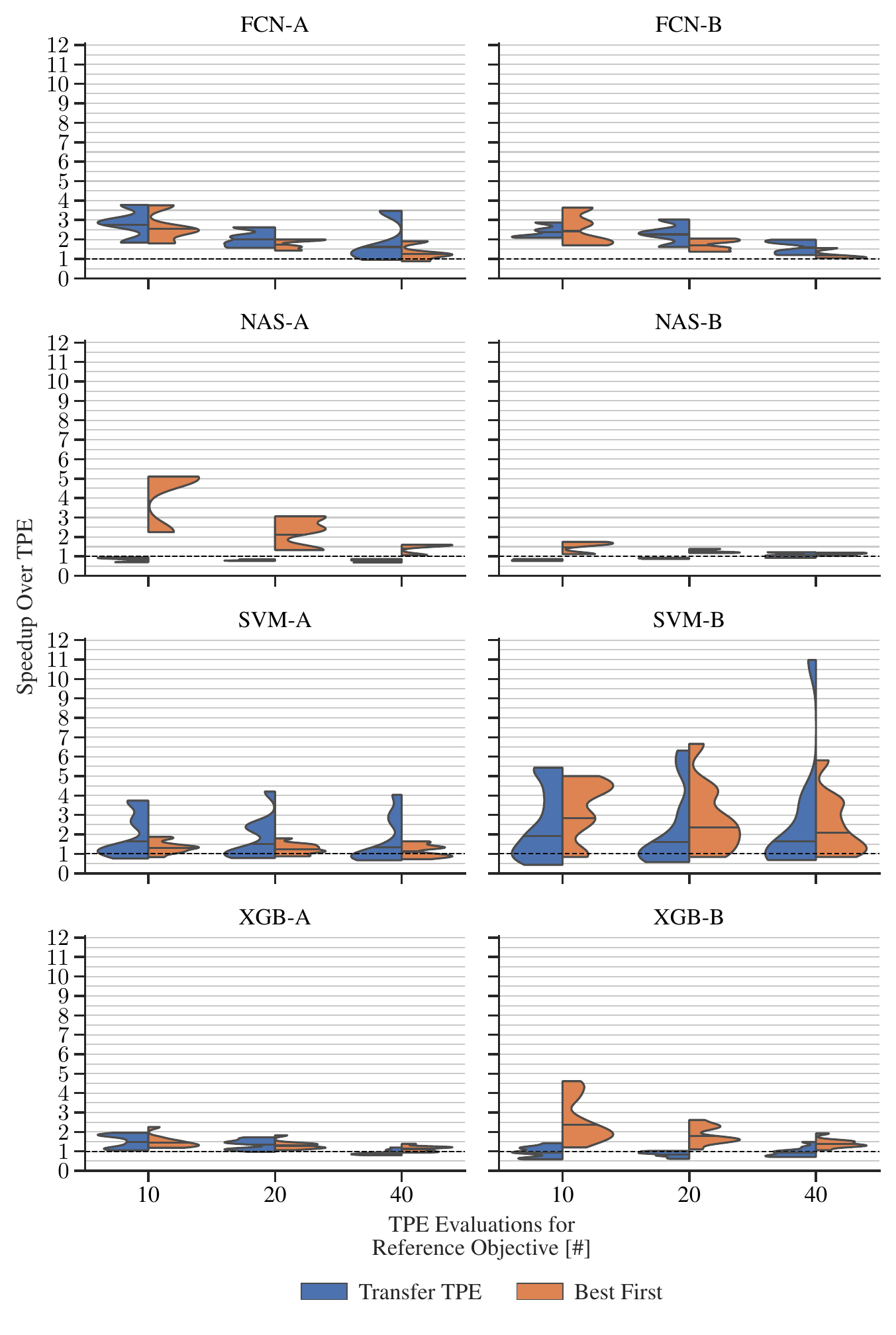}
    \vspace{-15pt}
    \caption{Speedup of transfer TPE and best-first over TPE across tasks for each of 8 benchmarks. The previous HPO has a budget of 20 evaluations. The violins estimate densities of the task geometric means. The horizontal line in each violin shows the geometric mean across these task means. The x-axis shows the budget for the TPE reference.}
    \label{fig:speedup_transfer_tpe_no_best_first+transfer_tpe_no_ttpe_over_tpe_eval_ref_20}
\end{figure}

\begin{figure}[H]
    \centering
    \includegraphics{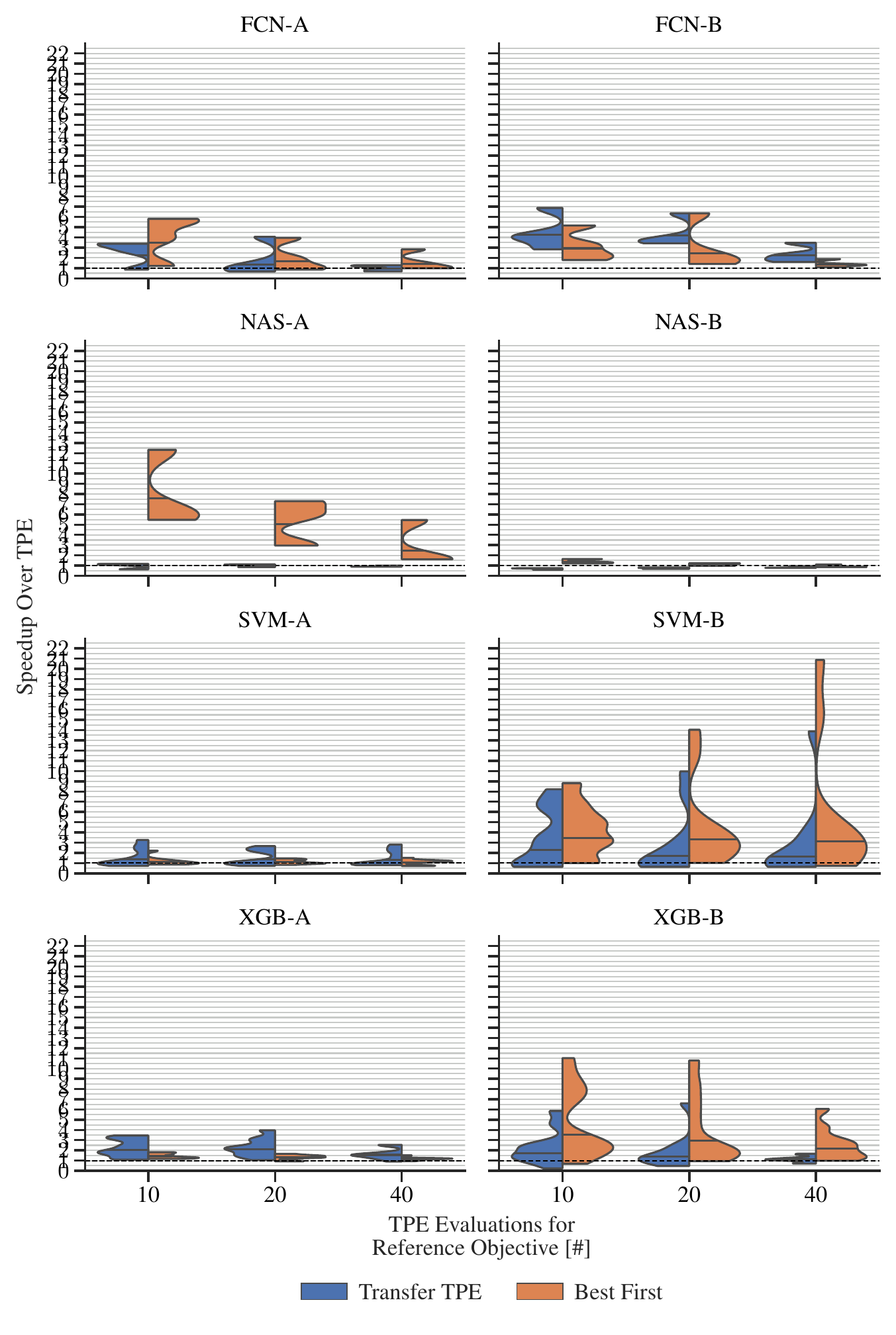}
    \vspace{-15pt}
    \caption{Speedup of transfer TPE and best-first over TPE across tasks for each of 8 benchmarks. The previous HPO has a budget of 40 evaluations. The violins estimate densities of the task geometric means. The horizontal line in each violin shows the geometric mean across these task means. The x-axis shows the budget for the TPE reference.}
    \label{fig:speedup_transfer_tpe_no_best_first+transfer_tpe_no_ttpe_over_tpe_eval_ref_40}
\end{figure}

\section{Speedup Combined Best First and Transfer TPE}\label{apx:combined_speedup}

\begin{figure}[H]
    \centering
    \includegraphics{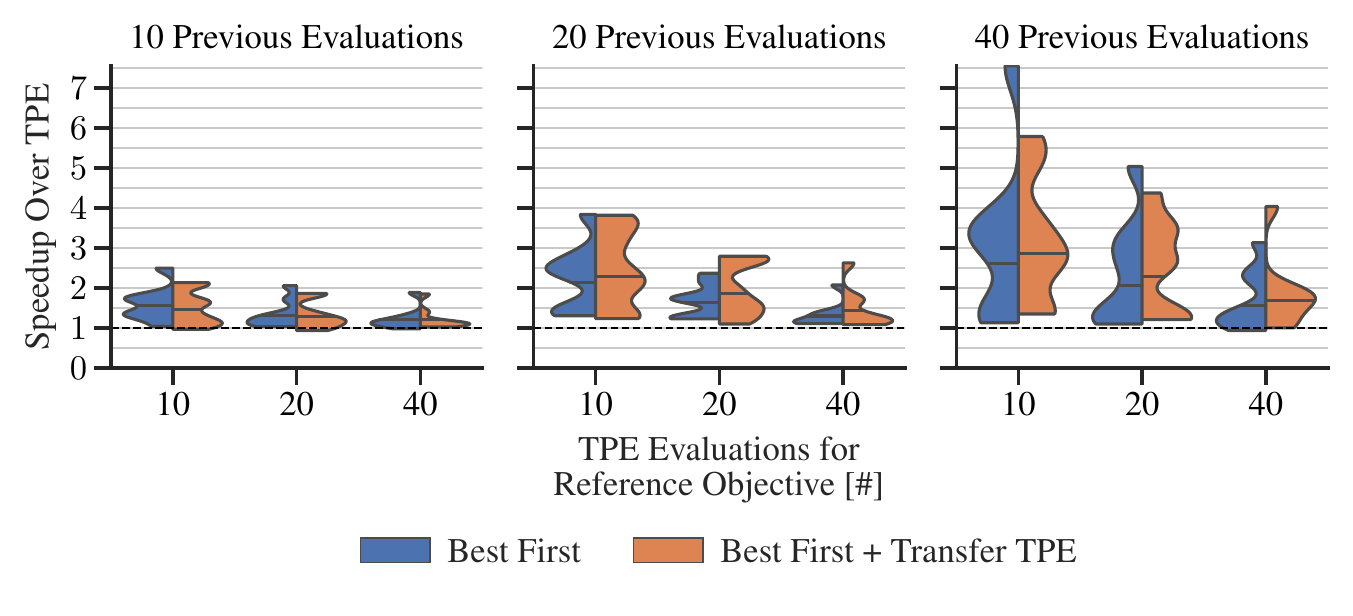}
    \vspace{-15pt}
    \caption{Speedup to reach a given reference objective value compared to TPE for best-first and combined best-first with transfer TPE across 8 benchmarks. The violins estimate densities of benchmark geometric means. The horizontal line in each violin shows the geometric mean across these benchmark means. \#Evaluations for the old HPO increases from left to right. The x-axis shows the budget for the TPE reference.}
    \label{fig:global_speedup_transfer_tpe_no_ttpe+transfer_tpe_over_tpe}
\end{figure}

\section{Failure Rates}\label{apx:main_failure}

\begin{figure}[H]
    \centering
    \includegraphics{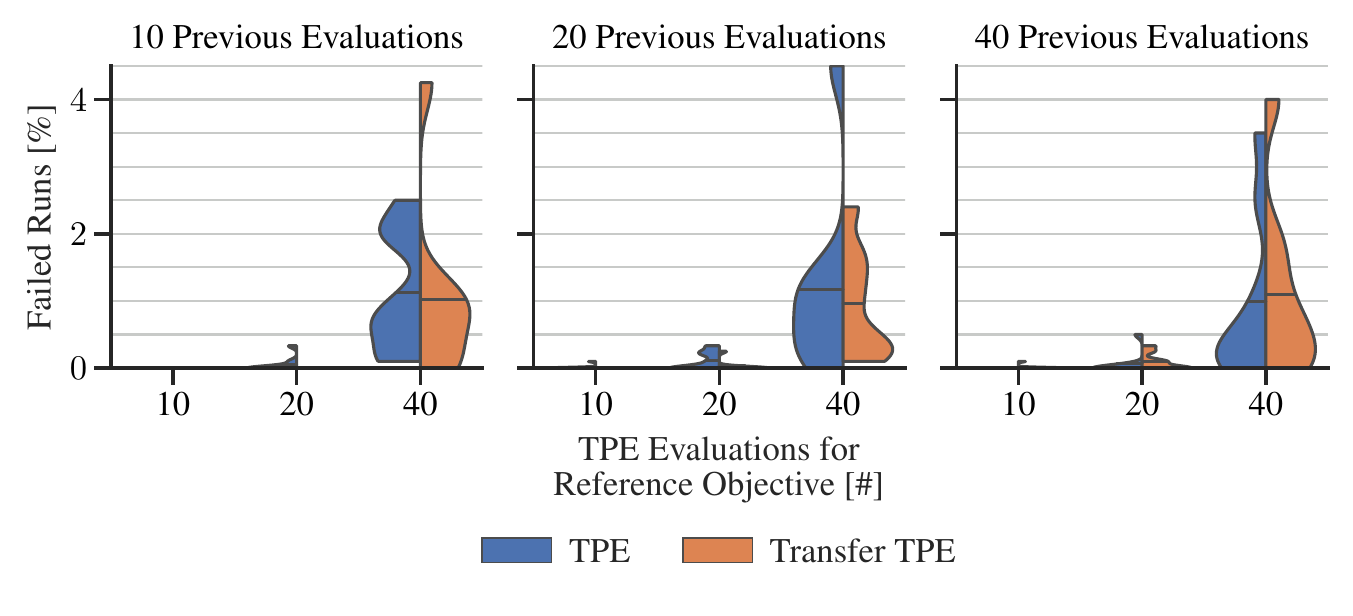}
    \vspace{-15pt}
    \caption{Failure rates for transfer TPE and TPE across 8 benchmarks. The violins estimate densities of the task means. The horizontal line in each violin shows the mean across these task means. The plots from left to right utilize increasing budget for the pre-adjustment hyperparameter. The x-axis shows the budget for the TPE reference.}
    \label{fig:global_nan_percent_transfer_tpe_no_best_first_and_tpe}
\end{figure}

\begin{figure}[H]
    \centering
    \includegraphics{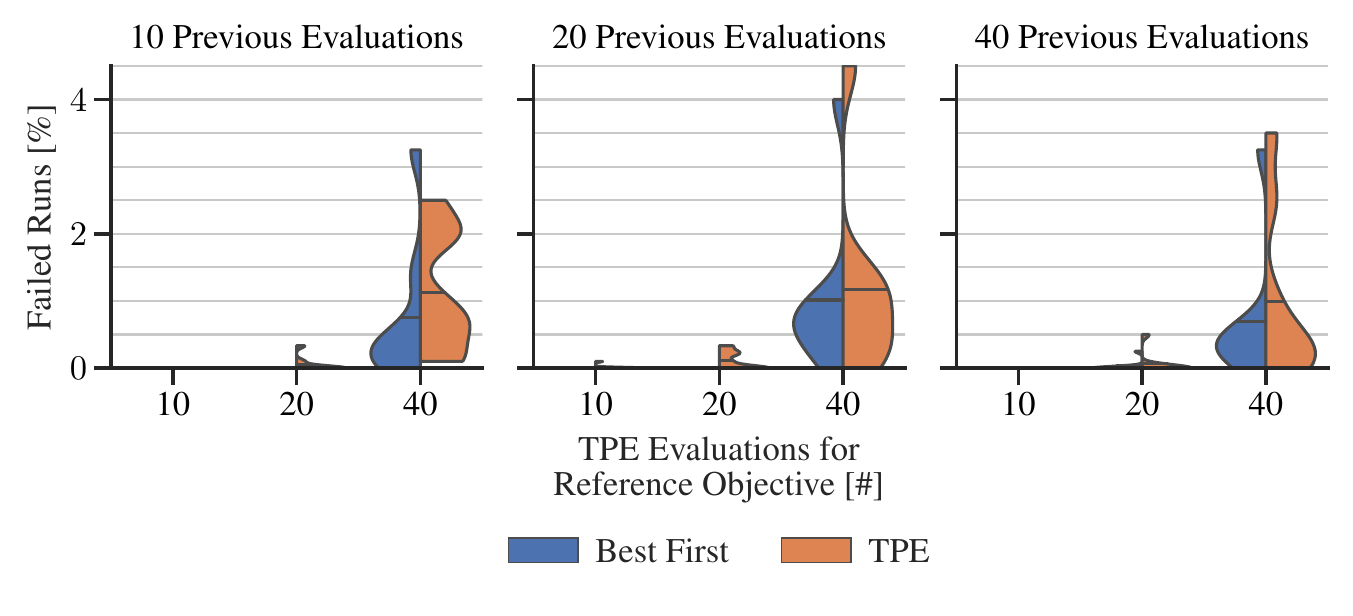}
    \vspace{-15pt}
    \caption{Failure rates for best-first and TPE across 8 benchmarks. The violins estimate densities of task means. The horizontal line in each violin shows the mean across these task means. \#Evaluations for the old HPO increases from left to right. The x-axis shows the budget for the TPE reference.}
    \label{fig:global_nan_percent_transfer_tpe_no_ttpe_and_tpe}
\end{figure}

\section{Objective Improvements}\label{apx:improvement_analysis}

For the improvement plots, we show the difference of means normalized with respect to the standard deviation of the control algorithm.
This metric is known as glass delta.
As some benchmarks had a standard deviation of 0, we added a small constant in those cases.
We chose this constant according to the 0.2-quantile of the observed values.
For the plots we clip the improvement to $[-100, \, \infty)$, as for some plots there are extreme outliers.

\subsection{Transfer TPE and Best First vs. TPE}

\begin{figure}[H]
    \centering
    \includegraphics{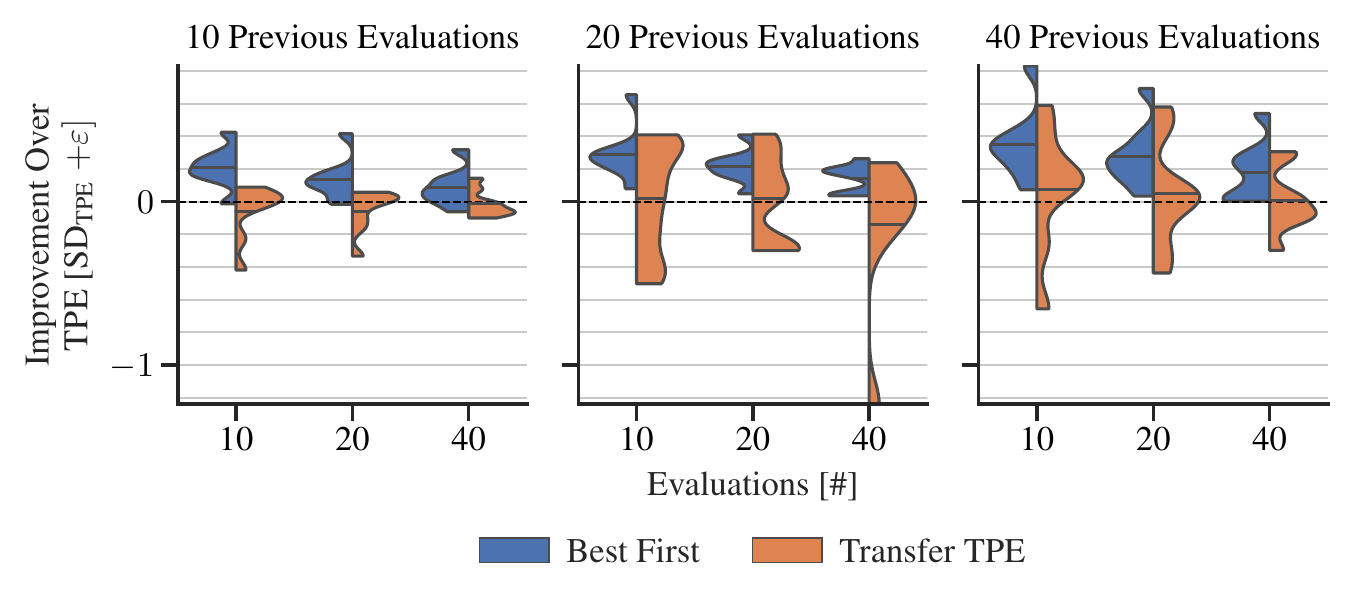}
    \vspace{-15pt}
    \caption{Standardized objective improvements of Transfer TPE and best-first over TPE across 8 benchmarks. The violins estimate densities of the benchmark means. The horizontal line in each violin shows the mean across these benchmark means. \#Evaluations for the old HPO increases from left to right. In each plot, the evaluation budget increases.}
    \label{fig:global_improvement_transfer_tpe_no_best_first+transfer_tpe_no_ttpe}
\end{figure}

\subsection{Only Optimize New and Drop Unimportant vs. TPE}

\begin{figure}[H]
    \centering
    \includegraphics{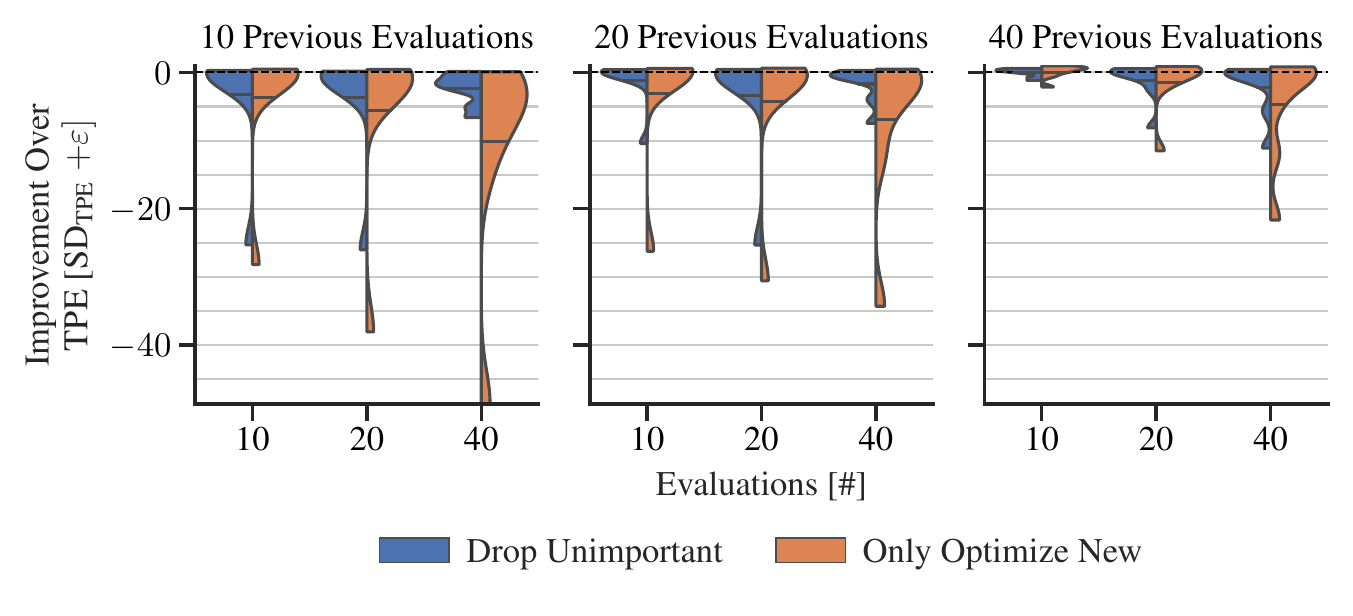}
    \vspace{-15pt}
    \caption{Standardized objective improvements of only-optimize-new and drop-unimportant over TPE across 8 benchmarks. The violins estimate densities of the benchmark means. The horizontal line in each violin shows the mean across these benchmark means. \#Evaluations for the old HPO increases from left to right. In each plot, the evaluation budget increases.}
    \label{fig:global_improvement_transfer_top+transfer_importance}
\end{figure}

\section{Control Study: TPE for Different Random Seed Ranges}\label{apx:control_study_tpe2}
\begin{figure}[H]
    \centering
    \includegraphics{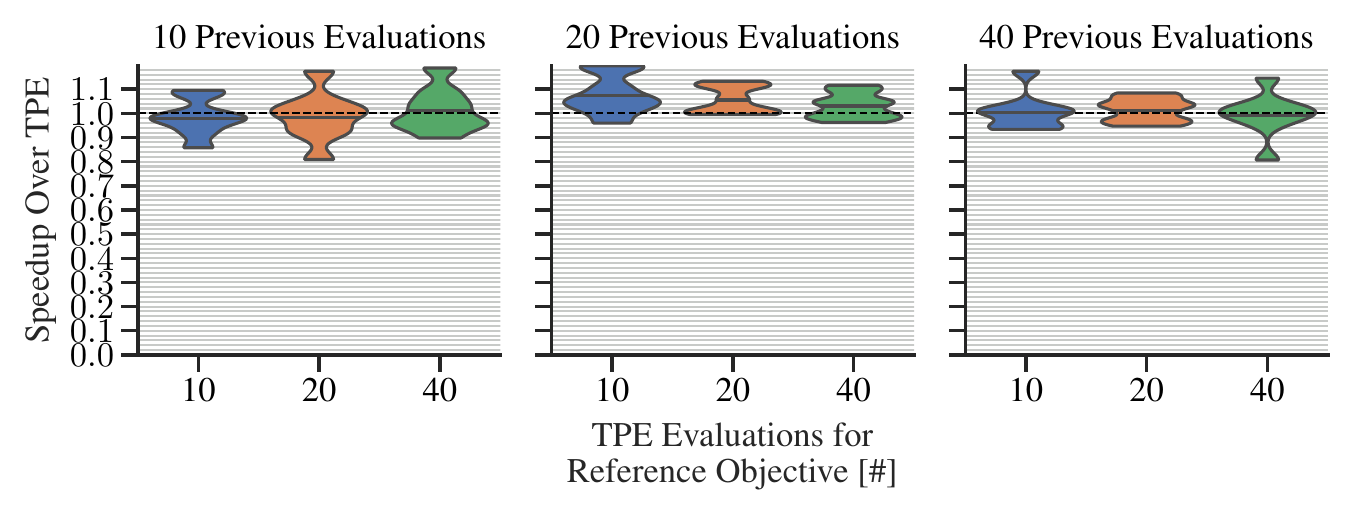}
    \vspace{-5pt}
    \caption{Speedup of TPE over TPE2 across 8 benchmarks. The violins estimate densities of the benchmark geometric means. The horizontal line in each violin shows the geometric mean across these benchmark means. \#Evaluations for the old HPO increases from left to right. The x-axis shows the budget for the TPE reference.}
    \label{fig:global_speedup_tpe2_over_tpe}
\end{figure}

As a sanity check, and to gauge the influence of random seeds, we compare TPE to itself with different seed ranges.
In general we observe little differences in TPE and TPE2, with the exception of one outlier task (\Figref{fig:global_speedup_tpe2_over_tpe}).

\section{Control Study: Random Search vs TPE}\label{apx:control_study_random}
\begin{figure}[H]
    \centering
    \includegraphics{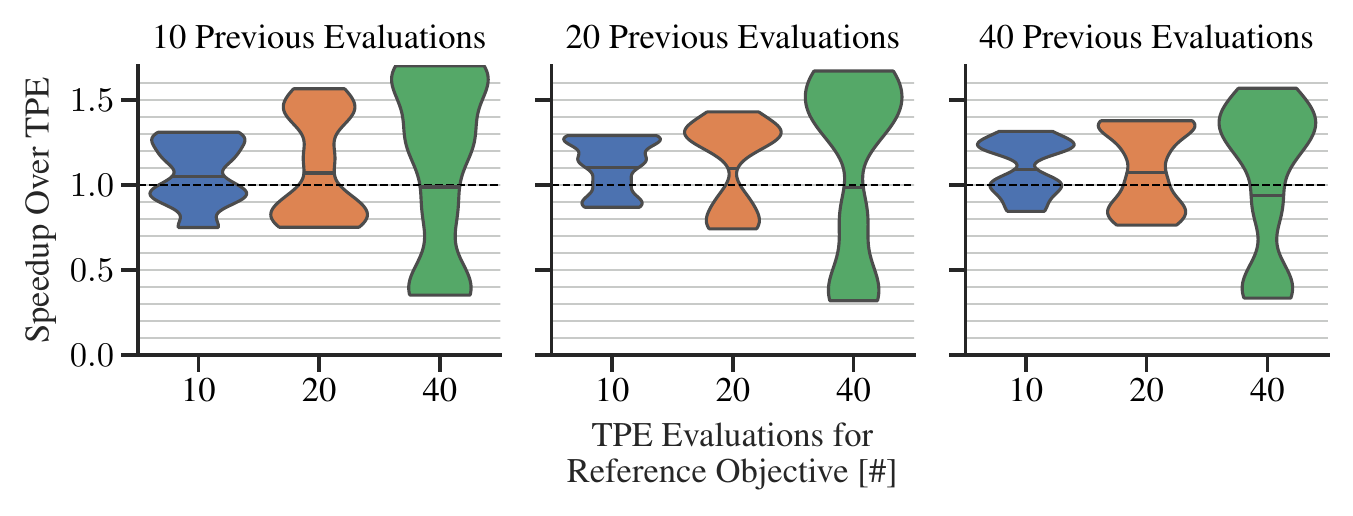}
    \vspace{-5pt}
    \caption{Speedup of random search over TPE across 8 benchmarks. The violins estimate densities of the benchmark means. The horizontal line in each violin shows the geometric mean across these benchmark means. \#Evaluations for the old HPO increases from left to right. The x-axis shows the budget for the TPE reference.}
    \label{fig:global_speedup_random_over_tpe}
\end{figure}

As a sanity check, and for context, we compare TPE to random search (\Figref{fig:global_speedup_random_over_tpe}).

\end{document}